\def\year{2021}\relax
\documentclass[letterpaper]{article} 
\usepackage{aaai21}  
\usepackage{times}  
\usepackage{helvet} 
\usepackage{courier}  
\usepackage[hyphens]{url}  
\usepackage{graphicx} 
\usepackage{natbib}  
\usepackage{caption} 
\usepackage{epsfig}
\usepackage{amsmath}
\usepackage{amssymb}
\usepackage{adjustbox}
\usepackage{bm}
\usepackage{multirow}
\usepackage{booktabs}
\usepackage{color}
\usepackage{mathrsfs}
\usepackage{cite}
\usepackage{url}
\usepackage[switch]{lineno}
\usepackage[ruled, vlined, linesnumbered]{algorithm2e}

\title{SSPC-Net: Semi-supervised Semantic 3D Point Cloud Segmentation Network}
\author {
	Mingmei Cheng,
	Le Hui,
	Jin Xie\thanks{Corresponding author.},
	Jian Yang\\
}
\affiliations {
	PCA Lab, Key Lab of Intelligent Perception and Systems for High-Dimensional Information of Ministry of Education\\
	Jiangsu Key Lab of Image and Video Understanding for Social Security\\
	School of Computer Science and Engineering, Nanjing University of Science and Technology, Nanjing, China\\
	\{chengmm, le.hui, csjxie, csjyang\}@njust.edu.cn
}
\begin{document}

\maketitle

\begin{abstract}
	Point cloud semantic segmentation is a crucial task in 3D scene understanding. Existing methods mainly focus on employing a large number of annotated labels for supervised semantic segmentation. Nonetheless, manually labeling such large point clouds for the supervised segmentation task is time-consuming. In order to reduce the number of annotated labels, we propose a semi-supervised semantic point cloud segmentation network, named SSPC-Net, where we train the semantic segmentation network by inferring the labels of unlabeled points from the few annotated 3D points.   
	In our method, we first partition the whole point cloud into superpoints and build superpoint graphs to mine the long-range dependencies in point clouds. Based on the constructed superpoint graph, we then develop a dynamic label propagation method to generate the pseudo labels for the unsupervised superpoints. Particularly, we adopt a superpoint dropout strategy to dynamically select the generated pseudo labels. In order to fully exploit the generated pseudo labels of the unsupervised superpoints, we furthermore propose a coupled attention mechanism for superpoint feature embedding. Finally, we employ the cross-entropy loss to train the semantic segmentation network with the labels of the supervised superpoints and the pseudo labels of the unsupervised superpoints. Experiments on various datasets demonstrate that our semi-supervised segmentation method can achieve better performance than the current semi-supervised segmentation method with fewer annotated 3D points.
\end{abstract}

\section{Introduction}

Due to the increasing growth of 3D point cloud data, point cloud semantic segmentation has been receiving more and more attention in the 3D computer vision community. Most of these segmentation methods focus on fully supervised segmentation with manually annotated points \cite{hu2020randla, thomas2019kpconv, lei2020seggcn, zhao2019pointweb, wang2019graph}. However, annotating large-scale 3D point clouds is a cumbersome process, which is costly in labor and time. Particularly, the number of point clouds in some real scenes such as the indoor scene can often reach the order of magnitude to millions. Therefore, it is difficult to obtain the accurate labels of these million points for full-supervised segmentation.

Different from full-supervised point cloud segmentation, semi-supervised segmentation aims to learn a good label prediction for point clouds with partially annotated points. Recent works have been dedicated to the semi-supervised point cloud segmentation task.
Guinard~\emph{et al.}~\cite{guinard2017weakly} propose a weakly supervised conditional random field classifier for 3D LiDAR point cloud segmentation.
However, it converts the segmentation task into an optimization problem, and the contextual information in point clouds is ignored. Mei~\emph{et al.} propose a semi-supervised 3D LiDAR point cloud segmentation method~\cite{mei2019semantic}, where the 3D data is projected to range images for feature embedding, and the inter-frame constraints are combined with some labeled samples to encourage feature consistency. Nonetheless, the constraints along the LiDAR sequential frames are not available in general 3D segmentation datasets. Lately,~\cite{XuLee_CVPR20} proposes a semi-supervised point cloud segmentation method, which employs three constraints to enhance the feature learning of unlabeled points, including block-level label penalization, data augmentation with rotation and flipping for prediction consistency, and a spatial and color smoothness constraint in local regions. Although it can obtain effective segmentation results, the long-range relations are ignored in this method.

Although some efforts have been made on semi-supervised point cloud segmentation, how to accurately predict the labels of unannotated points for segmentation is still a challenging problem. Particularly, since point clouds are irregular, it is difficult to exploit the geometry structures of point clouds to accurately infer pseudo labels of unannotated points for label propagation. In addition, the uncertainty of inferred pseudo labels of unannotated points hinders the network from learning discriminative features of point clouds, leading to inaccurate label prediction.

Aiming at the aforementioned two problems, in this paper, we propose a novel semi-supervised semantic point cloud segmentation network, named SSPC-Net. We first divide the point clouds into superpoints and build the superpoint graph, where the superpoint is a set of points with isotropically geometric features. Thus, we can convert the point-level label prediction problem in the point cloud segmentation task into the superpoint-level label prediction problem. Following the method in ~\cite{landrieu2018large}, we employ the gated graph neural network (GNN)~\cite{li2015gated} for superpoint feature embedding. 
In order to fully exploit the local geometry structure of the constructed superpoint graph, we then develop a dynamic label propagation method to accurately infer pseudo labels for unsupervised superpoints. Specifically, the labels of supervised superpoints are gradually extended to the adjacent superpoints with high semantic similarity along the edges of the superpoint graph. 
We also adopt a superpoint dropout strategy to obtain the high-quality pseudo labels during the label propagation process, where the extended superpoints with low confidences are dynamically pruned.
Furthermore, we propose a coupled attention mechanism to learn the discriminative context features of superpoints. We alternatively perform attention on the supervised and extended superpoints so that the discrimination of the features of the supervised and extended superpoints can be boosted each other, alleviating the uncertainty of the inferred pseudo labels of the unsupervised superpoints.
Finally, we employ a combined cross-entropy loss to train the segmentation network.
Extensive results on various indoor and outdoor datasets demonstrate that our method can yield good performance with only few point-level annotations.

The main contributions of this paper are summarized as: \textbf{(1)} We develop a dynamic superpoint label propagation method to accurately infer the pseudo labels of unsupervised superpoints. We also present a superpoint dropout strategy to select the high-quality pseudo labels. \textbf{(2)} We propose a coupled attention mechanism on the supervised and extended superpoints to learn the discriminative features of the superpoints. \textbf{(3)} Our proposed method can yield better performance than the current semi-supervised point cloud semantic segmentation method with fewer labels.

\section{Related Work}
\textbf{Deep learning on 3D point clouds.}
Recently, many deep learning methods are proposed to tackle point cloud classification and segmentation.
Some methods~\cite{wu20153d,maturana2015voxnet,sedaghat2016,qi2016volumetric} voxelize point clouds and employ 3D CNNs for feature embedding. However, the voxel-based methods suffer from the large memory cost due to the high-resolution voxels. By projecting point clouds into 2D views, \cite{su15mvcnn,boulch2017unstructured,tatarchenko2018tangent} use classic CNNs to extract features from point clouds. However, the view-based methods are sensitive to the density of 3D data.
To reduce memory cost and additional preprocessing, Qi~\emph{et al.} propose PointNet, which directly processes the unordered point clouds and uses multi-layer perceptrons (MLPs) and the maxpooling function for feature embedding. Following PointNet, many efforts~\cite{qi2017pointnet++,klokov2017escape,wang2019graph,hua2018pointwise,li2018pointcnn,zhao2019pointweb,wang2019dynamic,thomas2019kpconv,wu2019point,liu2019point2sequence,han2020point2node,zhao2020jsnet,feng2018gvcnn,ma2018learning} are proposed for point cloud processing. Although these methods have achieved decent performance, their models depend on fully annotated 3D point clouds for training. 
However, in this paper, we focus on the semi-supervised point cloud semantic segmentation.

\textbf{Semi-/Weakly supervised deep learning on 3D point clouds.}
Many efforts~\cite{mei2019semantic,wei2020multi,XuLee_CVPR20} have been proposed to tackle semi-/weakly supervised point cloud semantic segmentation. In \cite{mei2019semantic}, Mei~\emph{et al.} introduce a semi-supervised 3D LiDAR data segmentation method. It first converts the 3D data to depth maps and then applies CNNs for feature embedding. In addition to a small part of supervised data, it also leverages the temporal constraints along the LiDAR scans sequence to boost feature consistency. Therefore, it is not practicable for general point cloud segmentation cases. Inspired by CAM~\cite{zhou2016learning}, Wei~\emph{et al.} propose MPRM~\cite{wei2020multi} with scene-level and subcloud-level labels for weakly supervised segmentation. Specifically, it leverages a point class activation map (PCAM) to obtain the localization of each class and then generates point-wise pseudo labels with a multi-path region mining module. In this way, the segmentation network can be trained in a fully supervised manner. However, in practice, generating the subcloud-level annotation is still time-consuming. 
Lately, in~\cite{XuLee_CVPR20}, Xu~\emph{et al.} propose a semi-supervised algorithm, which uses three constraints on the unlabeled points, $i.e.$, the block level labels for penalizing the negative categories in point clouds, data augmentation with random in-plane rotation and flipping for feature consistency and a spatial and color smoothness constraint in point clouds. 

\begin{figure*}[t]
	\begin{center}
		\includegraphics[width=0.9\linewidth]{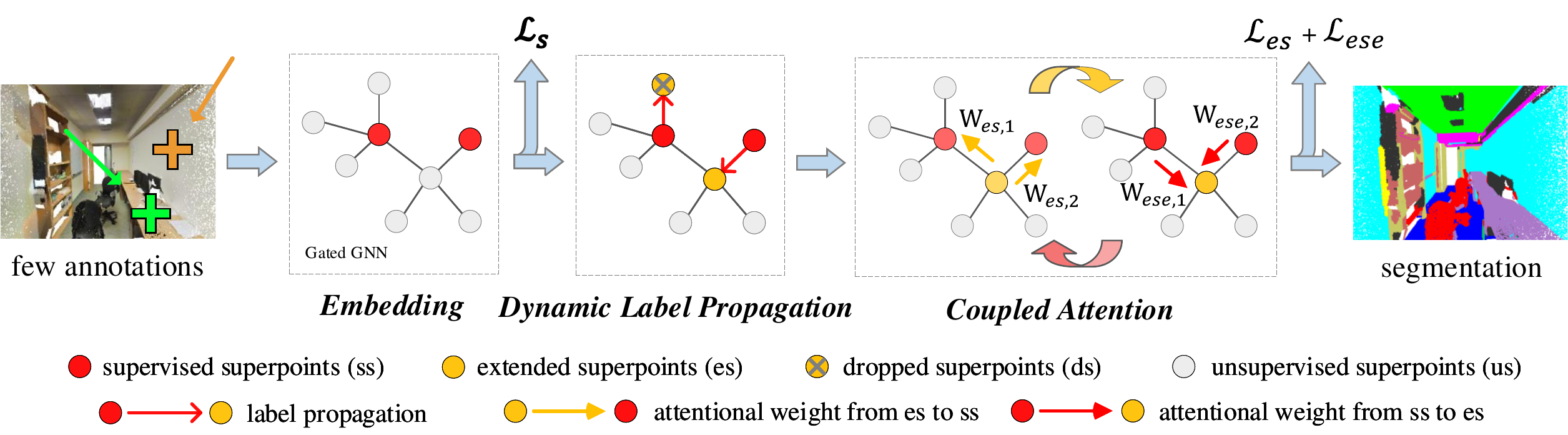}
	\end{center}
	\caption{Overview of the proposed semi-supervised semantic point cloud segmentation network (SSPC-Net). We first leverage the gated GNN to extract superpoints features. Then based on the superpoint graph, we conduct the dynamic label propagation strategy to generate pseudo labels. Next, based on the supervised superpoints and the extended superpoints, we perform a coupled attention mechanism to further boost the extraction of discriminative contextual features in the point cloud.}
	\label{fig_outline}
\end{figure*}

\section{Our Method}
In this section, we present our semi-supervised point cloud segmentation network and the outline of our framework is shown in Fig. \ref{fig_outline}. We first introduce the superpoint graph embedding module. Then we propose a dynamic label propagation approach combined with a superpoint dropout strategy. Next, we propose a coupled attention mechanism to learn discriminative contextual features of superpoints. Finally, we depict the framework of our method.

\subsection{Superpoint Graph Embedding}\label{sec_graph_embeddding}

To obtain the superpoints and learn the superpoint features, following \cite{landrieu2018large}, we perform an unsupervised superpoints partition approach to generate superpoints and then build superpoint graphs combined with graph neural network (GNN) for superpoints feature embedding.
Denote $\mathcal{G=(V, E)}$ as the superpoint graph built upon superpoints, where $\mathcal{V}$ is the node set and $\mathcal{E}$ is the edge set. Edge $(i,j) \in \mathcal{E}$ links node $i \in \mathcal{V}$ with $j \in \mathcal{V}$. 
We first perform a lightweight PointNet-like structure on the superpoints to obtain superpoints features. After that, we learn the superpoint embedding with the gated GNN used in~\cite{li2015gated}.
Given the superpoint embeddings and the semi-supervision, we can penalize the model with incomplete supervision. For a point cloud consists of $N$ superpoints, we define $a_i \in \{0,1\}^{N}$ to indicate whether the $i$-th superpoint has supervision. Then the segmentation loss $\mathcal{L}_{s}$ on the superpoint graph embedding module can be formulated as:
\begin{equation}
\mathcal{L}_{s} = \frac{1}{A} \sum \nolimits _{i=1}^{N}a_i \cdot \mathcal{F}_{loss}\left(z_i, \bm{y}_i\right)
\end{equation}
where $\mathcal{F}_{loss}$ is the loss function and we choose the cross-entropy loss in experiments, $A=\sum \nolimits _{i=1}^{N} a_i$ is adopted for normalization, $z_i$ represents the superpoint-level label of $i$-superpoint and $\bm{y}_i$ is the prediction logit.

The reason why we choose the superpoint graph as the representation of point cloud is at two points. On the one hand, the superpoint is geometrically isotropic and therefore we can directly extend the point-level label to the superpoint-level label, which alleviates the lack of supervision. On the other hand, since the superpoint graph is rooted in the geometric structure of the point cloud, where the linking edges between the superpoints greatly facilitate the feature propagation. Thus we can obtain more discriminative contextual features of superpoints.

\begin{figure}[t]
	\begin{center}
		\includegraphics[width=0.95\linewidth]{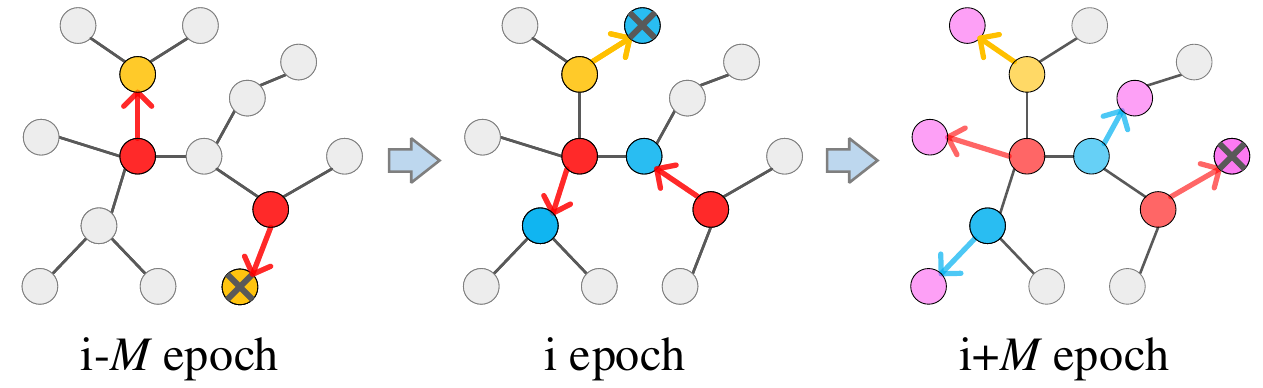}
	\end{center}
	\caption{The procedure of our dynamic label propagation. We progressively propagate the superpoint-level label and discard the extended superpoint with low confidence.}
	\label{fig_extension}
\end{figure}

\subsection{Dynamic Label Propagation}\label{sec_extension_growing}
To propagate superpoint labels, we propose a dynamic label propagation strategy to generate pseudo labels. 
Suppose we have constructed three sets: the supervised superpoints set $S$, unsupervised superpoints set $U$, and extended superpoints set $E$. Note that at the beginning we set $E=\varnothing$. Besides, elements in each set indicate the index of superpoints.

For $\forall ~i \in T, T=S\cup E$, we use the adjacent superpoints to construct candidate set $\mathcal{N}_i$, where we consider propagating labels in it. Suppose $z_i$ is the label of $i$-th superpoint. Note that $\forall ~j \in\mathcal{N}_i$ must satisfy two constraints: $j\in U$ and the predicted category of the $j$-th superpoint should be the same as that of the $i$-th superpoint, that is, $z_i$. Compared with other unsupervised superpoints, elements in $\mathcal{N}_i$ are  with higher possibilities to be assigned with pseudo labels, due to the close geometric relations and the close distances to the $i$-th superpoint. To generate high-quality pseudo labels, we assess the confidence scores of the superpoints in $\mathcal{N}_i$ and denote the scores as $\bm{m}_i\in\mathbb{R}^{|\mathcal{N}_i|}$. Then, we enumerate all the superpoints in $\mathcal{N}_i$ and select the superpoint with the highest confidence score. The operation can be formulated as:
\begin{equation}
j^*=\mathop{\arg\max}\limits_{j=1,2,\ldots, |\mathcal{N}_i|}({m}_{i,j})
\label{eqn:constraint}
\end{equation}
where $j^*$ represents the index of the superpoint with the highest confidence score in $\mathcal{N}_i$. To further ensure the high quality of pseudo labels, we set the threshold $\tau$ to filter the selected superpoints with dissatisfactory confidence values. When the confidence score $m_{i,j^*}\geqslant \tau$, the $j^*$-th superpoint is selected and assigned with pseudo label $z_i$. Then $j^*$ will be removed from the unsupervised superpoints set $U$ and added to the extended superpoints set $E$. On the contrary, if there is no superpoint satisfying the constraint, no superpoint will be extended from $\mathcal{N}_{i}$. In the experiments, $\tau$ is empirically set to 0.9. Note that for each extension procedure, we merge the supervised superpoints set $S$ and the extended superpoints set $E$ to the new set $T=S\cup E$ for further extension. Because the extended superpoints with pseudo labels can also be treated as the superpoints with supervision for further label propagation. In this way, we can progressively propagate the labels of the supervised superpoints and generate more high-quality pseudo labels for unsupervised superpoints in $U$. What's more, Algorithm \ref{algorithm_extension} shows the details of the graph-based supervision extension procedure.

\begin{algorithm}[t]
	\DontPrintSemicolon
	\KwIn{Supervised superpoints set $S$, unsupervised superpoints set $U$, extended superpoints set $E$, threshold $\tau$}
	\KwOut{Updated sets $U$ and $E$}
	$T = S\cup E$ \\
	\For {$~i \in T$}
	{
		Denote $z_i$ as the label of $i$-th superpoint \\
		Construct the candidate supeproints set $\mathcal{N}_i$ \\
		\If{$\mathcal{N}_i\neq \varnothing $}
		{
			Generate the confidence scores $\bm{m}_i$ \\
			$j^*=\mathop{\arg\max}\limits_{j=1,2,\ldots, |\mathcal{N}_i|}({m}_{i,j})$ \\
			\If{$m_i^{j^*}\geqslant \tau$}
			{
				Assign pseudo label $z_i$ to the $j^*$-th superpoint \\
				$U := U \setminus \{j^*\} \quad E := E \cup \{j^*\}$
			}
		}
		
	}
	\caption{Graph-based supervision extension}\label{algorithm_extension}
\end{algorithm}

Since our extension strategy is performed progressively, we consider removing the low-confidence superpoints in the extended superpoints set $E$. Hence, we propose a superpoint dropout strategy assessing the reliability of the extended superpoints in the embedding space. In the superpoints set $T=S\cup E$, we cluster the superpoints into $c$ classes according to the superpoints labels or pseudo labels, where $c$ is the number of categories. Suppose $\mathcal{C}_i$ is the $i$-th cluster set that contains the index of the superpoints belonging to the $i$-th category. In addition, we denote $\bm{v}_i$ as the feature of the cluster center of $\mathcal{C}_i$, which is computed by averaging the features of all the superpoints in $\mathcal{C}_i$. We assess the confidence of the extended superpoints by considering its distance to the corresponding cluster center in the feature space. For $\forall ~j\in E\cap \mathcal{C}_i$, its Euclidean distance to the cluster center in the feature space is formulated as:  
\begin{equation}
d_i^j = \left\|\bm{f}_{j}-\bm{v}_i \right \|_2
\end{equation} 
where $\bm{f}_j\in\mathbb{R}^{D}$ is the feature of $j$-th superpoint, and $\bm{v}_i\in\mathbb{R}^{D}$ is the feature of cluster center. Smaller distance indicates the higher reliability of extended superpoints, whereas the larger distance means the higher uncertainty. Therefore, in each cluster, we discard $k$ extended superpoints that are furthest from the cluster center, where $k$ is set to 0.05$*|E\cap\mathcal{C}_i|$. In other words, we retain the most reliable 95\% superpoints and drop the 5\% unreliable superpoints in the set $E\cap\mathcal{C}_i$. Our superpoint dropout strategy is explained in Algorithm \ref{algorithm_dropout}.

Concretely, as shown in Fig. \ref{fig_extension}, we perform our graph-based dynamic label propagation strategy every ${M}$ epochs, therefore, the extended superpoints are gradually ``growing'' on the graph from the supervised superpoints. The reason why we conduct the extension operation in a multi-stage manner instead of every epoch is that our extension strategy is a cumulative one, which means that too much extension operations will cause redundant extended superpoints and aggravate the memory cost. Meanwhile, the model is not stable at the beginning, which is not conducive to generating extended superpoints. 

\begin{algorithm}[t]
	\DontPrintSemicolon
	\KwIn{Number of classes $c$, supervised superpoints set $S$, unsupervised superpoints set $U$, extended superpoints set $E$}
	\KwOut{Updated sets $U$ and $E$}
	$T = S\cup E$ \\
	Cluster on $T$ and obtain $c$ cluster sets: $\mathcal{C}_1,\mathcal{C}_2, \ldots,\mathcal{C}_c$ \\
	\For {$i = 1:c $}
	{			
		Compute the feature $\bm{v}_i$ of the cluster center of $\mathcal{C}_i$ \\
		\For{each $j \in E\cap\mathcal{C}_i$}
		{
			Generate the feature $\bm{f}_j$ of the $j$-th superpoint \\
			Compute the distance $d_i^j = \left\|\bm{f}_{j}-\bm{v}_i \right \|_2$
		}
		Find the farthest 5\% superpoints (set as $\mathcal{C}_{drop}$) in $E\cap\mathcal{C}_i$ from the cluster center according to the distance $\bm{d}_i$ \\
		$E := E \setminus \mathcal{C}_{drop}\quad U := U \cup \mathcal{C}_{drop}$
	}
	\caption{Superpoint dropout strategy}\label{algorithm_dropout}
\end{algorithm}

\subsection{Coupled Attention for Feature Enhancement}\label{sec_feature_enhance}
Aiming to learn more discriminative contextual features in point clouds, we propose a coupled attention mechanism.
For $\forall ~i\in S$, we denote the corresponding embedding as $\bm{h}_i \in \mathbb{R}^{D}$. Similarly, for $\forall ~j\in E$, we denote the corresponding embedding as $\bm{h}_j$. By weighing all the extended superpoints, we extract the novel contextual feature of $i$-th superpoint with attention mechanism:
\begin{equation}
\bm{x}_i=\sum \nolimits_{j\in E} g\left(\phi(\bm{h}_i, \bm{h}_j) \right)\odot \alpha(\bm{h}_j)
\end{equation}
where $\phi(\bm{h}_i, \bm{h}_j)=\mathop{MLP}(\bm{h}_i-\bm{h}_j)$ embeds the channel-wise relations between superpoints, $\alpha(\bm{h}_j)=\mathop{MLP}(\bm{h}_j)$ is a unary function for individual superpoint embedding, $\phi(\cdot,\cdot):\mathbb{R}^{D} \rightarrow \mathbb{R}^{D}$ and $\alpha:\mathbb{R}^{D} \rightarrow \mathbb{R}^{D}$, $\odot$ is the Hadamard product. $g$ is a normalization function and is defined as:
\begin{equation}
g\left(\phi_l(\bm{h}_i, \bm{h}_j) \right) = \frac{\exp(\phi_{l}({\bm h}_{i}, {\bm h}_{j}))}{\sum\nolimits_{r \in E}\exp(\phi_{l}({\bm h}_{i}, {\bm h}_{r}))}
\end{equation}
where $l=1,2,\ldots,D$, represents $l$-th element of embedding $\phi_{l}(\cdot,\cdot)$. Consequently, the matrix representation of the attention operation on the supervised superpoints in $S$ can be formulated as:
\begin{equation}
\bm{X}_s = \sum \nolimits_{j\in E}\bm{W}_{es,j} \odot \bm{H}_{e,j}
\end{equation}
where $\bm{X}_s\in\mathbb{R}^{|S|\times D}$, $\bm{W}_{es,j}\in\mathbb{R}^{|S|\times D}$, $\bm{H}_{e,j} \in \mathbb{R}^{|S|\times D}$ and $j$ enumerates the extended superpoints in $E$. Note that $\bm{W}_{es}\in\mathbb{R}^{|S|\times|E|\times D}$ represents the channel-wise weights from the extended superpoints to the supervised superpoints.

Once we obtain the attention embedding $\bm{X}_s\in\mathbb{R}^{|S|\times D}$, we can derive new segmentation logits of supervised superpoints and formulate the loss as:
\begin{equation}
\mathcal{L}_{es} = \frac{1}{|S|} \sum \nolimits _{i \in S}\mathcal{F}_{loss}\left(z_i, \mathop{FC} \left(\bm{X}_{s,i}\right)\right)
\end{equation}
where $z_i$ is the superpoint-level label, $\bm{X}_{s,i}$ is the attention feature of the corresponding supervised superpoint, and $\mathcal{F}_{loss}$ is the cross-entropy loss adopted in experiments. Note that $\mathop
{FC}$ is the fully connected layer, which maps $\bm{X}_{s,i}\in\mathbb{R}^{|S|\times D}$ from the $D$-dim to the dimension of the categories.

\begin{figure}[t]
	\begin{center}
		\includegraphics[width=0.95\linewidth]{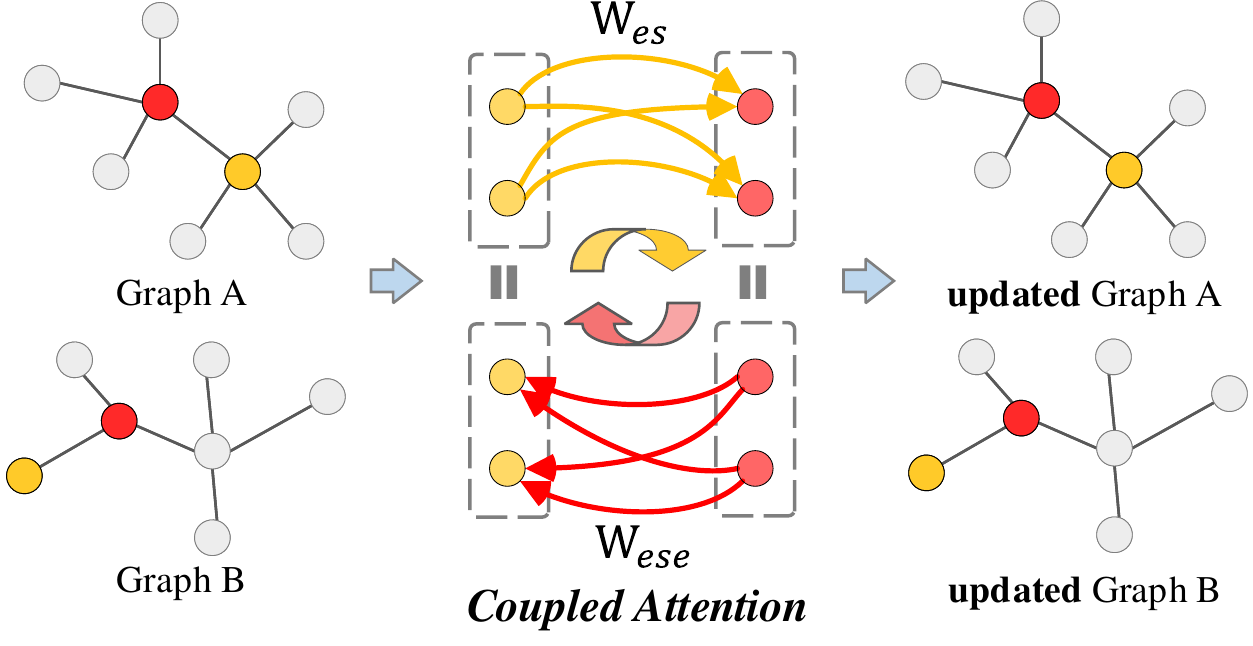}
	\end{center}
	\caption{The coupled attention for feature enhancement. 
	}
	\label{fig_attention}
\end{figure}

Similarly, to promote the feature characterization of the extended superpoints, we then perform attention on the extended superpoints in reverse. By weighting the new features enhanced by the attention operation of the supervised superpoints, we boost the context feature propagation and thus enhance the robustness of the features of the extended superpoints. Thus, for $\forall j \in E$, the new embedding of the corresponding superpoint can be calculated as:
\begin{equation}
\bm{y}_j=\sum \nolimits_{i\in S} g\left(\psi(\bm{h}_j, \bm{x}_i) \right)\odot \beta(\bm{x}_i)
\end{equation}
where $\psi(\bm{h}_j, \bm{x}_i)=\mathop{MLP}(\bm{h}_j-\bm{x}_i)$ characterizes the dependencies of the extended superpoints on the attention embeddings of the supervised superpoints. $\beta(\bm{x}_i)=\mathop{MLP}(\bm{x}_i)$ is a unary function similar to $\alpha$, $\psi(\cdot,\cdot):\mathbb{R}^{D} \rightarrow \mathbb{R}^{D}$ and $\beta:\mathbb{R}^{D} \rightarrow \mathbb{R}^{D}$. $g$ is a normalization function defined as:
\begin{equation}
g\left(\psi_{l}(\bm{h}_j, \bm{x}_i) \right) = \frac{\exp(\psi_{l}({\bm h}_{j}, {\bm x}_{i}))}{\sum\nolimits_{r \in S}\exp(\psi_{l}({\bm h}_{j}, {\bm x}_{r}))}
\end{equation}
where $l=1,2,\ldots,D$, denotes $l$-th element of embedding $\psi(\cdot,\cdot)$. Then the matrix representation of the attention operation on the extended superpoints in $E$ can be defined as:
\begin{equation}
\bm{Y}_e = \sum \nolimits_{i\in S}\bm{W}_{ese,i} \odot \bm{\mathcal{X}}_{s,i}
\end{equation}
where $\bm{Y}_e\in\mathbb{R}^{|E|\times D}$, $\bm{W}_{ese,i}\in\mathbb{R}^{|E|\times D}$, $\bm{\mathcal{X}}_{s,i}\in\mathbb{R}^{|E|\times D}$ and $i$ enumerates superpoints in $S$. Note that $\bm{\mathcal{X}}_{s}$ is the feature after employing function $\beta(\cdot)$ on attention feature $\bm{X}_{s}$. In this way, we develop the coupled attention, $i.e.$, $\bm{W}_{ese}\in\mathbb{R}^{|S|\times|E|\times D}$ denotes the channel-wise weights from the attentional supervised superpoints to extended superpoints.

Then the loss $\mathcal{L}_{ese}$ on the extended superpoints with enhanced attention features can be formulated as:
\begin{equation}
\mathcal{L}_{ese} = \frac{1}{|E|} \sum \nolimits _{j \in E}\mathcal{F}_{loss}\left(z^{p}_j, \mathop{FC} \left(\bm{Y}_{e,j}\right)\right)
\end{equation}
where $z^{p}_j$ is the pseudo label and $\mathcal{F}_{loss}$ is the cross-entropy loss as well. $\mathop{FC}$ maps the feature to the category space.

Specifically, as shown in Fig. \ref{fig_attention}, our coupled attention considers the intra- and inter-relations concurrently. To encourage the feature consistency in different point clouds, 
we integrate the supervised superpoints and extended superpoints in various point clouds into sets $S$ and $E$, respectively.
The connections between $S$ and $E$ are constructed within and across various point cloud samples, and superpoints with the same labels are encouraged to have more similar semantic embeddings compared to those with diverse classes. 
As a result, by alternatively performing attention on the supervised and extended superpoints, more long-range dependencies between superpoints are built. Hence, the model learns more discriminative and robust contextual features of the supervised and unsupervised superpoints.

\subsection{Framework}\label{sec_framework}
The framework of our model is illustrated in Fig. \ref{fig_outline}. In our framework, the superpoint graph embedding module is the basis of our point cloud feature embedding. Based on this module, the dynamic label propagation method assesses the semantic similarity between the superpoints and propagates the superpoint-level supervision along the edges of the superpoint graph. Then, with the extended superpoints searched by the dynamic label propagation module, we propose a coupled attention mechanism to boost the contextual feature learning of the point cloud. 

The final objective function is a combination of the three objectives $\mathcal{L}_{final}=\mathcal{L}_{s} + \lambda_1\cdot \mathcal{L}_{es} +\lambda_2\cdot \mathcal{L}_{ese}$ and we empirically set $\lambda_1, \lambda_2$ to 1.
As shown in Fig. \ref{fig_outline}, the dynamic label propagation module and coupled attention module are only conducted in the training stage. For testing, we obtain the inferred prediction directly from the superpoint graph embedding module.

\section{Experiments}
\subsection{Implementation Details}
To train our model, we adopt Adam optimizer with a base learning rate of 0.01. For the S3DIS~\cite{armeni20163d}, ScanNet~\cite{dai2017scannet} and vKITTI~\cite{Gaidon2016Virtual} dataset, we employ the mini-batch size of 4, 8, 8, respectively. We empirically implement the dynamic label propagation module every $M=40$ epochs.

\textbf{Semi-supervision generation.}
To produce the semi-supervision of point clouds, we randomly select a part of the points with annotations in each class. For example, given a point cloud containing $n$ points with $c$ classes, suppose the supervision rate be $r$, then we evenly distribute the supervision budget $r\cdot n$ and randomly sample $ (r\cdot n)/c$ points in each category as the supervised points.
The label of superpoint is the category with the most annotated points.
If there is no supervised point contained, then the superpoint will be unsupervised. 
Note that compared with the sampling strategy of random sampling annotated points directly in point clouds, our labeling mechanism is more in coincident with the human annotation behavior, since the random sampling strategy will result in that most of the supervised points will be occupied by the areas with simple geometric structure but more points, $e.g.$, walls, roads, etc.
For evaluation, all the quantitative results are computed at the point level.


\begin{table}[t]
	\centering
	\begin{adjustbox}{width=0.88\linewidth}
		\large
		\begin{tabular}{c|l|c|ccc} 
			\toprule
			\multicolumn{2}{c|}{{\bf Method}} &\multicolumn{1}{c|}{Rate}&mIoU&mAcc&OA\\
			\midrule
			\multicolumn{6}{c}{6-fold cross validation} \\
			\midrule
			\multirow{7}{*}{Full}
			&PointNet&100\%&47.6&66.2&78.5\\			
			&SPGraph &100\%& 62.1 &73.0 &85.5 \\
			&PointCNN &100\%&{\bf65.3} &{\bf75.6} &{\bf88.1} \\
			&RSNet&100\%&56.4&66.4&-\\
			& G+RCU2 &100\%&49.7 &66.4 &81.1 \\
			& 3P-RNN &100\%&56.3 &73.6 &86.9 \\
			\midrule
			\multirow{3}{*}{Semi-}
			&Baseline &0.002\% &45.1 &63.7 &73.9 \\
			&{\bf SSPC-Net} &0.002\% &48.5 &68.3 &79.1\\	
			&{\bf SSPC-Net} &0.01\%&{\bf54.5}&{\bf70.8} &{\bf 80.4}\\		
			\midrule
			\multicolumn{6}{c}{Fold 5} \\
			\midrule
			\multirow{5}{*}{Full}
			&PointNet&100\%&41.1&49.0&-\\
			&PointNet++&100\%&47.8&-&-\\
			&SPGraph &100\%&{\bf58.0} &{\bf66.5} &{\bf86.3} \\
			&SegCloud&100\%&48.9&57.3&-\\
			&PointCNN&100\%&57.2&63.8&85.9\\
			\midrule
			\multirow{6}{*}{Semi-}
			&Semi-Seg&1pt &44.5&-&-\\
			&Semi-Seg&10\% &48.0&-&-\\			
			&Baseline &0.002\% &39.6 &52.1 &72.4 \\	 	
			&{\bf SSPC-Net}&0.002\% &43.0 &56.4 &76.2\\
			&{\bf SSPC-Net}&0.01\%&51.5 &63.8 &82.0\\
			&{\bf SSPC-Net}&1pt &{\bf53.8} &{\bf63.9} &{\bf83.8}\\
			\bottomrule
		\end{tabular}
	\end{adjustbox}
	\caption{Evaluation on the S3DIS dataset.}
	\label{tab_results_s3dis}
\end{table}

\subsection{Semi-supervised Semantic Segmentation}
\textbf{S3DIS.}
S3DIS~\cite{armeni20163d} dataset is an indoor 3D dataset including 6 areas and 13 categories. Three metrics are adopted for quantitative evaluation: mean IoU (mIoU), mean class accuracy (mAcc), and overall accuracy (OA).

\begin{figure}[t]
	\begin{center}
		\includegraphics[width=0.87\linewidth]{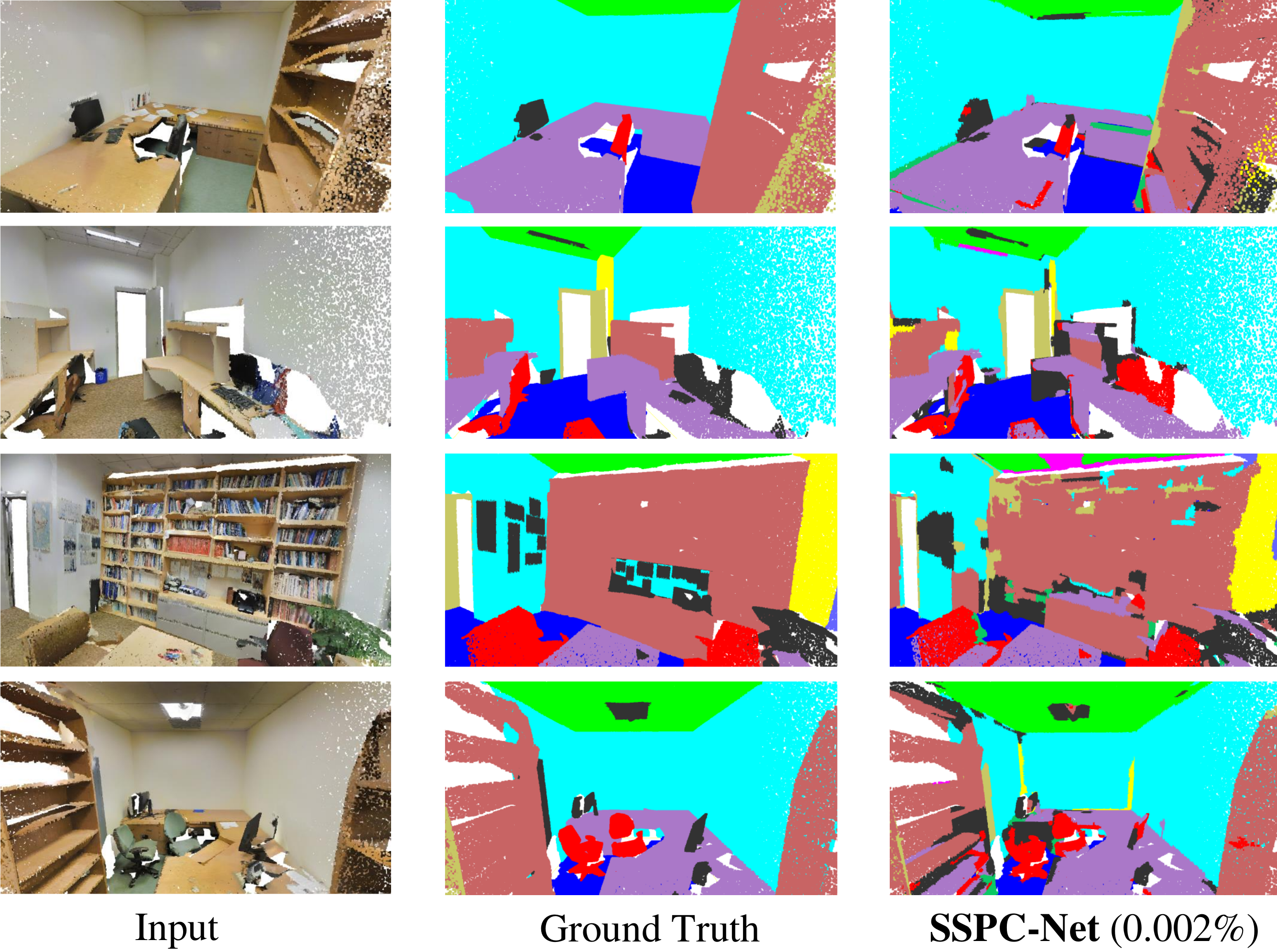}
	\end{center}
	\caption{The visual results on the S3DIS dataset with supervision rate of 0.002\%.}
	\label{fig_visual_results}
\end{figure}

The quantitative and visual results are shown in Tab. \ref{tab_results_s3dis} and Fig. \ref{fig_visual_results}, respectively. For a fair comparison, we test our framework with the ``1pt'' labeling strategy adopted in \cite{XuLee_CVPR20} (dubbed ``Semi-Seg'' in Tab. \ref{tab_results_s3dis}) as well, which samples one point in each category of each block as the supervised point. It can be seen that our SSPC-Net achieves a significant gain of 9.3\% in terms of mIoU with the ``1pt'' labeling strategy. In \cite{XuLee_CVPR20}, Xu {\em et~al.} split the point cloud into blocks and then train and test their model on each block separately. Nonetheless, our model learns the embeddings of superpoints in the whole point cloud, therefore we can obtain more discriminative contextual features and yield better performance. Note that in Tab.~\ref{tab_results_s3dis}, ``Baseline'' represents our method without the label propagation strategy and coupled attention mechanism. One can see that our SSPC-Net improves the performance from 39.6\% to 43.0\% in terms of mIoU with the supervision rate of 0.002\% on Area 5 of the S3DIS dataset, benefiting from the pseudo labels generated from the label propagation and the discriminative contextual features extracted by the coupled attention mechanism.

\textbf{ScanNet.}
ScanNet~\cite{dai2017scannet} is an indoor scene dataset containing 1513 point clouds with 20 categories. We split the dataset into a training set with 1201 scenes and a testing set with 312 scenes following \cite{qi2017pointnet++}. We adopt overall semantic voxel labeling accuracy (OA) and mean IoU (mIoU) for evaluation.

We list the quantitative results on the testing set in Tab.~\ref{tab_scannet_vkitti}. Similar to S3DIS, ScanNet is also an indoor dataset, but the point cloud of ScanNet is much sparser than that of S3DIS. This brings greater challenges to the propagation of supervised labels. However, the proposed model can still achieve good segmentation results and even outperform some fully supervised methods like PointNet \cite{qi2017pointnet} with semi-supervision. Furthermore, the performance of the proposed model is much better than the baseline method, which further validates the effectiveness of our method.

\begin{table}[t]
	\centering
	\LARGE
	\begin{adjustbox}{width=0.99\linewidth}
		\begin{tabular}{c|l|c|cc|ccc}
			\toprule
			\multicolumn{2}{c|}{\multirow{2}{*}{{\bf Method}}} & \multicolumn{1}{c|}{\multirow{2}{*}{Rate}} &
			\multicolumn{2}{c|}{ScanNet} & \multicolumn{3}{c}{vKITTI}  \\
			\multicolumn{2}{c|}{}                       & & mIoU & OA                    & mIoU & mAcc & OA            \\ 
			\midrule
			\multirow{7}{*}{Full}
			&PointNet&100\%	&-&73.9&34.4&47.0&79.7\\
			&PointNet++&100\%&-&{\bf84.5}  &-&-&-\\
			
			&SSP + SPG&100\% &- &- &{\bf52.0} &{\bf67.3} &84.3    \\ 
			&G+RCU&100\% &-&-      &35.6&57.6&79.7\\
			&RSNet&100\% &{\bf39.3} &79.2 &-&-&-    \\ 
			&3P-RNN&100\%&-&- &41.6&54.1&{\bf87.8} \\
			&3DCNN&100\%&-&73.0 &-&-&- \\
			\midrule
			\multirow{3}{*}{Semi-}
			&Baseline&0.01\% &24.1&38.2 &35.7 &53.4 &79.2 \\
			&{\bf SSPC-Net}&0.01\% &27.1&66.6 &41.0 &55.7 &81.2 \\
			&{\bf SSPC-Net}&0.05\% &{\bf39.3}&{\bf77.1} &{\bf50.6} &{\bf64.8} &{\bf85.4} \\
			\bottomrule
		\end{tabular}
	\end{adjustbox}
	\caption{Evaluation on the ScanNet and vKITTI datasets.}
	\label{tab_scannet_vkitti} 	 
\end{table}

\textbf{vKITTI.}
vKITTI~\cite{Gaidon2016Virtual} dataset mimics the real-world KITTI dataset and contains the synthetic outdoor scenes with 13 classes (including road, tree, terrain, car, etc.). For evaluation, we split the dataset into 6 non-overlapping sub-sequences and employ 6-fold cross validation following \cite{ye20183d}. Mean IoU (mIoU), mean class accuracy (mAcc)
and overall accuracy (OA) are employed for evaluation.

The quantitative results are presented in Tab.~\ref{tab_scannet_vkitti}. With the 0.01\% point-level annotations, compared with the baseline method, our model achieves better segmentation results due to the dynamic label propagation strategy and the discriminative contextual features generated from the coupled attention module. In addition, our model can achieve better or comparable performance than some fully supervised methods with only 0.01\% and 0.05\% of the supervised points.

\subsection{Ablation Study}
\textbf{Contribution of individual components.}
In this section, we investigate the contribution of the proposed components to model performance. The evaluation results on Area5 of the S3DIS dataset of different components with the supervision ratio of 0.002\% and 0.01\% are shown in Tab. \ref{tab_components}, where the components are the graph embedding (Graph Emb.), dynamic label propagation (Label Prop.), coupled attention for feature enhancement (Coup. Attn.). It can be observed that there is an obvious promotion on the performance with the addition of dynamic label propagation and coupled attention module, which further demonstrates the effectiveness of these strategies for the semi-supervision.

\begin{table}[t]
	\centering
	\huge
	\begin{adjustbox}{width=1.0\linewidth}
		\begin{tabular}{ccc|ccc|ccc}
			\toprule
			\multicolumn{3}{c|}{{\bf Components}} & \multicolumn{3}{c|}{Rate$=$0.002\%}                                           & \multicolumn{3}{c}{Rate$=$0.01\%}                                            \\
			\midrule
			Graph     & Label    & Coup.    & \multirow{2}{*}{mIoU} & \multirow{2}{*}{mAcc} & \multirow{2}{*}{OA} & \multirow{2}{*}{mIoU} & \multirow{2}{*}{mAcc} & \multirow{2}{*}{OA} \\
			Emb. & Prop. & Attn. & & & & &  &   \\		
			\midrule
			\checkmark	&			&			&39.6 &52.1 &72.4 &48.5 &61.2 &80.3 \\ 
			\checkmark	&\checkmark &	        &40.9 &55.8 &73.6 &50.0 &60.6 &80.8 \\ 
			\checkmark	&\checkmark	&\checkmark	&{\bf 43.0} &{\bf 56.4}  &{\bf 76.2} &{\bf 51.5} &{\bf 63.8}  &{\bf 82.0}\\
			\bottomrule
		\end{tabular}
	\end{adjustbox}
	\caption{The contribution of different components on Area5 of the S3DIS dataset with different annotation rates.}
	\label{tab_components}	
\end{table}

\begin{figure}[t]
	\centering
	\includegraphics[width=0.78\linewidth]{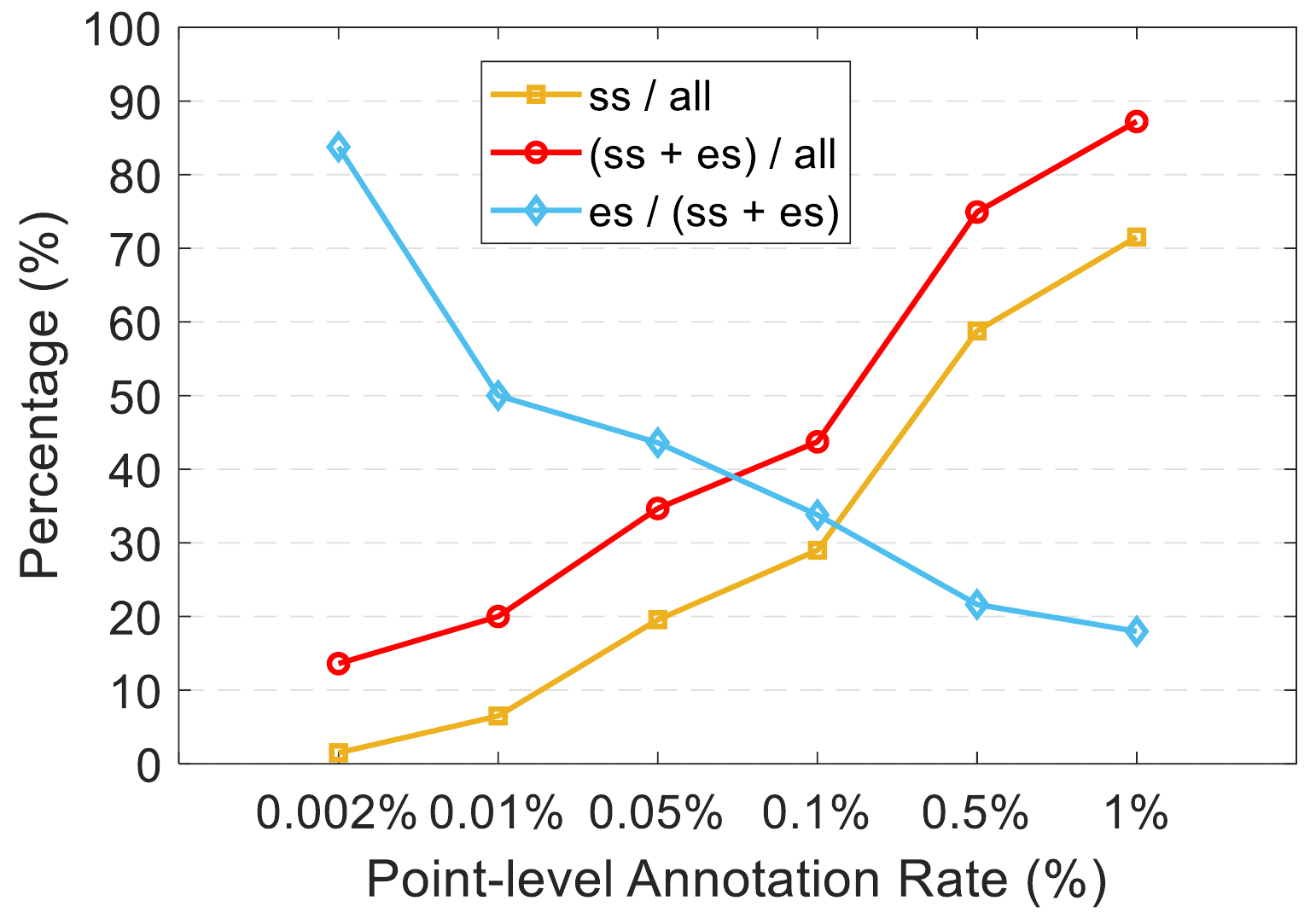}
	\caption{The percentage of supervised superpoints (ss) and extended superpoints (es) during training. Note that ``all'' means the overall superpoints.}
	\label{fig_ext_ablation}
\end{figure}

\textbf{Supervision rate.} The number of supervised points plays an important role in the segmentation performance. The more labeled points, the smaller gap of data distribution between the semi-supervision and full supervision. To discuss the effect of various labeling rates on model performance, we test our method on Area5 of the S3DIS dataset. The results are shown in Tab. \ref{tab_comparison}. Combined with Tab. \ref{tab_results_s3dis}, it can be observed that with only few labeled points, our model has already achieved effective segmentation results. With the growth of supervision, the performance of our model further increases. It is worth noting that we pay more attention to the cases of extremely few supervision signals, which is more challenging for the point cloud segmentation task.

\begin{table}[t]
	\centering
	\large
	\begin{adjustbox}{width=0.83\linewidth}
		\begin{tabular}{c|ccc|c}
			\toprule
			\;\;\;\;\;{\bf Rate}\;\;\;\;\;&\;\;mIoU & mAcc&OA\;\;&\;OA of es\; \\
			\midrule
			0.002\%      &43.0 &56.4 &76.2     & 87.3 \\
			0.01\%      &51.5 &63.8 &82.0      &90.9 \\
			0.1\%       &56.2 &66.1 &84.6      & 91.0 \\
			1.0\%       &58.3 &66.5 &85.7      &90.1 \\
			\bottomrule
		\end{tabular}
	\end{adjustbox}
	\caption{Comparison of various supervision rates on Area5 of the S3DIS dataset, where ``es'' represents the extended superpoints.} 
	\label{tab_comparison}
\end{table}

\begin{table}[t]
	\centering
	\LARGE
	\begin{adjustbox}{width=0.82\linewidth}
		\begin{tabular}{c|ccc}
			\toprule
			\;\;\;\;{\bf Interval} \bm{$M$}\;\;\;\; &\;\;\;mIoU\;\;\; &\;\;\;mAcc\;\;\; &\;\;\;OA\;\;\; \\
			\midrule
			20 &50.2 &61.1 &81.2 \\
			30 &50.8 &63.3 &81.5 \\			
			40 &{\bf 51.5} &{\bf 63.8} &{\bf 82.0} \\
			50 &49.6 &61.5 &80.7 \\
			60 &49.9 &62.2 &81.0 \\	
			\bottomrule
		\end{tabular}
	\end{adjustbox}
	\caption{Comparison of segmentation results with various interval $M$ of the dynamic label propagation method in the case of the supervision rate of 0.01\%.} 
	\label{tab_extension_M}
\end{table}

\textbf{Number of the extended superpoints.} The dynamic label propagation strategy plays an important role in our model. As shown in Fig. \ref{fig_ext_ablation}, we show the proportion of the supervised superpoints and extended superpoints in the training set when testing on Area 5 of the S3DIS dataset. With the increase of the annotated points, the proportion of the supervised superpoints increases rapidly. Because the probability of a superpoint containing a supervised point is getting higher as well. However, when there are fewer supervised points, the percentage of extended superpoints is obviously larger.
This demonstrates the importance of pseudo labels facing extremely few point annotations.

\textbf{Quality of the extended superpoints.} To analyze the quality of the extended superpoints, we evaluate the overall accuracy of the extended superpoints (OA of es) in Tab. \ref{tab_comparison}. Noted that, similar to the aforementioned metrics, the quantitative results of the extended superpoints are conducted at the point level as well. 
From Tab. \ref{tab_comparison}, one can see that the overall accuracy of extended superpoints is around 90\%, which demonstrates the high quality of extended superpoints. This further proves the effectiveness of our label propagation strategy which generates high-quality pseudo labels. In addition, the high quality of pseudo labels of the extended superpoints further reveals the reason for the improved performance based on the label propagation module.

\textbf{Epoch interval in dynamic label propagation.}
During the training, we perform the dynamic label propagation method every $M$ epochs. For comparison, we train our model with various interval $M$ while keeping other parameters unchanged with the supervision rate of 0.01\%. The evaluation results on Area 5 of the S3DIS dataset are shown in Tab. \ref{tab_extension_M}. It can be observed that when $M=40$, our model achieves the best performance.

\section{Conclusion}
In this paper, we proposed a semi-supervised point cloud segmentation network. We first partitioned the point cloud into superpoints and built superpoint graphs to explore the long-range relations in the point cloud. Then based on superpoint graphs, we proposed a dynamic label propagation method combined with a superpoint dropout strategy to generate high-quality pseudo labels for the unsupervised superpoints. Next, we proposed a coupled attention module to learn discriminative contextual features of superpoints and fully exploit the generated pseudo labels. Our method can achieve better performance than the current semi-supervised point cloud segmentation methods with fewer labels.

\section{Acknowledgments}
This work was supported by the National Science Fund of China (Grant Nos.
U1713208, 61876084), Program for Changjiang Scholars.

{
	\bibliographystyle{IEEEtran}
	\bibliography{egbib}
}
\end{document}